\documentclass{article} 
\usepackage{iclr2022_conference,times}

\usepackage{amsmath,amsfonts,bm}

\def\eqref#1{equation~\ref{#1}}

\def\1{\bm{1}}

\DeclareMathAlphabet{\mathsfit}{\encodingdefault}{\sfdefault}{m}{sl}
\SetMathAlphabet{\mathsfit}{bold}{\encodingdefault}{\sfdefault}{bx}{n}

\usepackage{hyperref}
\usepackage{url}
\usepackage{booktabs}       
\usepackage{array}
\usepackage{multicol}

\usepackage{pythonhighlight}

\usepackage{algorithm}
\usepackage{algorithmicx}
\usepackage{algpseudocode}
\algnewcommand\algorithmicforeach{\textbf{for}}
\algdef{S}[FOR]{ForEach}[1]{\algorithmicforeach\ #1\ \algorithmicdo}
\algnewcommand\algorithmicassert{\textbf{Assert}}
\algnewcommand\Assert[1]{\State \algorithmicassert #1}%

\algnewcommand\algorithmicreturnx{\textbf{return}}
\algnewcommand\Returnx[1]{\State \algorithmicreturnx #1}%

\usepackage{rotating}
\usepackage{sidecap}
\sidecaptionvpos{figure}{c}

\title{The UniNAS framework: combining modules in arbitrarily complex configurations with argument trees}

\author{Kevin Alexander Laube\\
Department of Cognitive Systems\\
University of Tuebingen, Germany\\
{\tt\small kevin.laube@uni-tuebingen.de}
}

\usepackage{graphicx}
\usepackage{amsmath}
\usepackage{amssymb}
\usepackage{caption}
\usepackage{subcaption}
\usepackage{booktabs} 
\usepackage{multirow}
\usepackage{enumitem}

\def\dcifar100{CIFAR100}
\def\dcifar10{CIFAR10}

\def\bnb201{NAS-Bench-201}

\iclrfinalcopy 
\begin{document}

\maketitle

\begin{abstract}
	Designing code to be simplistic yet to offer choice is a tightrope walk.
	Additional modules such as optimizers and data sets make a framework useful to a broader audience, but the added complexity quickly becomes a problem.
	Framework parameters may apply only to some modules but not others, be mutually exclusive or depend on each other, often in unclear ways.
	Even so, many frameworks are limited to a few specific use cases.
	This paper presents the underlying concept of UniNAS, a framework designed to incorporate a variety of Neural Architecture Search approaches.
	Since they differ in the number of optimizers and networks, hyper-parameter optimization, network designs, candidate operations, and more, a traditional approach can not solve the task.
	Instead, every module defines its own hyper-parameters and a local tree structure of module requirements.
	A configuration file specifies which modules are used, their used parameters, and which other modules they use in turn.
	This concept of \textit{argument trees} enables combining and reusing modules in complex configurations while avoiding many problems mentioned above.
	\textit{Argument trees} can also be configured from a graphical user interface so that designing and changing experiments becomes possible without writing a single line of code.
	UniNAS is publicly available at \url{https://github.com/cogsys-tuebingen/uninas}
\end{abstract}

\section{Preface}
\label{s_preface}

This paper primarily serves as a reference for my Ph.D. dissertation, which I am currently writing.
As a consequence, the framework is not under active development.
The presented concepts, problems, and solutions may be interesting regardless, even for other problems than Neural Architecture Search (NAS).
The framework's name, UniNAS, is a wordplay of University and Unified NAS since the framework was intended to incorporate almost any architecture search approach.

\section{Introduction and Related Work}
\label{s_introduction}

An increasing supply and demand for automated machine learning causes the amount of published code to grow by the day. Although advantageous, the benefit of such is often impaired by many technical nitpicks.
This section lists common code bases and some of their disadvantages.

\subsection{Available NAS frameworks}
\label{u_introduction_available}

The landscape of NAS codebases is severely fragmented, owing to the vast differences between various NAS methods and the deep-learning libraries used to implement them.
Some of the best supported or most widely known ones are:

\begin{itemize}
\setlength{\itemsep}{0pt}
\setlength{\parskip}{0pt}
\setlength{\parsep}{0pt}
	\item {NASLib~\citep{naslib2020}}
	\item {
		Microsoft NNI \citep{ms_nni} and Archai \citep{ms_archai}
	}
	\item {
		Huawei Noah Vega \citep{vega}
	}
	\item {
		Google TuNAS \citep{google_tunas} and PyGlove \citep{pyglove} (closed source)
	}
\end{itemize}

Counterintuitively, the overwhelming majority of publicly available NAS code is not based on any such framework or service but simple and typical network training code.
Such code is generally quick to implement but lacks exact comparability, scalability, and configuration power, which may be a secondary concern for many researchers.
In addition, since the official code is often released late or never, and generally only in either TensorFlow~\citep{tensorflow2015-whitepaper} or PyTorch~\citep{pytorch},
popular methods are sometimes re-implemented by some third-party repositories.

Further projects include the newly available and closed-source cloud services by, e.g., Google\footnote{\url{https://cloud.google.com/automl/}}
and Microsoft\footnote{\url{https://www.microsoft.com/en-us/research/project/automl/}}. Since they require very little user knowledge in addition to the training data, they are excellent for deep learning in industrial environments.

\subsection{Common disadvantages of code bases}
\label{u_introduction_disadvantages}

With so many frameworks available, why start another one?
The development of UniNAS started in early 2020, before most of these frameworks arrived at their current feature availability or were even made public.
In addition, the frameworks rarely provide current state-of-the-art methods even now and sometimes lack the flexibility to include them easily.
Further problems that UniNAS aims to solve are detailed below:

\paragraph{Research code is rigid}

The majority of published NAS code is very simplistic.
While that is an advantage to extract important method-related details, the ability to reuse the available code in another context is severely impaired.
Almost all details are hard-coded, such as:
\begin{itemize}
\setlength{\itemsep}{0pt}
\setlength{\parskip}{0pt}
\setlength{\parsep}{0pt}
	\item {
		the used gradient optimizer and learning rate schedule
	}
	\item {
		the architecture search space, including candidate operations and network topology
	}
	\item {
		the data set and its augmentations
	}
	\item {
		weight initialization and regularization techniques
	}
	\item {
		the used hardware device(s) for training
	}
	\item {
		most hyper-parameters
	}
\end{itemize}
This inflexibility is sometimes accompanied by the redundancy of several code pieces that differ slightly for different experiments or phases in NAS methods.
Redundancy is a fine way to introduce subtle bugs or inconsistencies and also makes the code confusing to follow.
Hard-coded details are also easy to forget, which is especially crucial in research where reproducibility depends strongly on seemingly unimportant details.
Finally, if any of the hard-coded components is ever changed, such as the optimizer, configurations of previous experiments can become very misleading.
Their details are generally not part of the documented configuration (since they are hard-coded), so earlier results no longer make sense and become misleading.

\paragraph{A configuration clutter}

In contrast to such simplistic single-purpose code, frameworks usually offer a variety of optimizers, schedules, search spaces, and more to choose from.
By configuring the related hyper-parameters, an optimizer can be trivially and safely exchanged for another. Since doing so is a conscious and intended choice, it is also documented in the configuration. In contrast, the replacement of hard-coded classes was not intended when the code was initially written.
The disadvantage of this approach comes with the wealth of configurable hyper-parameters, in different ways:

Firstly, the parametrization is often cluttered.
While implementing more classes (such as optimizers or schedules) adds flexibility, the list of available hyper-parameters becomes increasingly bloated and opaque.
The wealth of parametrization is intimidating and impractical since it is often nontrivial to understand exactly which hyper-parameters are used and which are ineffective.
As an example, the widely used PyTorch Image Models framework~\citep{rw2019timm} (the example was chosen due to the popularity of the framework, it is no worse than others in this respect) implements an intimidating mix of regularization and data augmentation settings that are partially  exclusive.\footnote{\url{https://github.com/rwightman/pytorch-image-models/blob/ba65dfe2c6681404f35a9409f802aba2a226b761/train.py}, checked Dec. 1st 2021; see lines 177 and below.}

Secondly, to reduce the clutter, parameters can be used by multiple mutually exclusive choices.
In the case of the aforementioned PyTorch Image Models framework, one example would be the selection of gradient-descent optimizers.
Sharing common parameters such as the learning rate and the momentum generally works well, but can be confusing since, once again, finding out which parameters affect which modules necessitates reading the code or documentation.

Thirdly, even with an intimidating wealth of configuration choices, not every option is covered. To simplify and reduce the clutter, many settings of lesser importance always use a sensible default value.
If changing such a parameter becomes necessary, the framework configurations become more cluttered or changing the hard-coded default value again results in misleading configurations of previous experiments.

To summarize, the hyper-parametrization design of a framework can be a delicate decision, trying for them to be complete but not cluttered.
While both extremes appear to be mutually exclusive, they can be successfully united with the underlying configuration approach of UniNAS: argument trees.

\paragraph{}
Nonetheless, it is great if code is available at all.
Many methods are published without any code that enables verifying their training or search results, impairing their reproducibility.
Additionally, even if code is overly simplistic or accompanied by cluttered configurations, reading it is often the best way to clarify a method's exact workings and obtain detailed information about omitted hyper-parameter choices.

\section{Argument trees}
\label{u_argtrees}

The core design philosophy of UniNAS is built on so-called \textit{argument trees}.
This concept solves the problems of Section~\ref{u_introduction_disadvantages} while also providing immense configuration flexibility.
As its basis, we observe that any algorithm or code piece can be represented hierarchically.
For example, the task to train a network requires the network itself and a training loop, which may use callbacks and logging functions.

Sections~\ref{u_argtrees_modularity} and~\ref{u_argtrees_register} briefly explain two requirements: strict modularity and a global register.
As described in Section~\ref{u_argtrees_tree}, this allows each module to define which other types of modules are needed. In the previous example, a training loop may use callbacks and logging functions.
Sections~\ref{u_argtrees_config} and~\ref{u_argtrees_build} explain how a configuration file can fully detail these relationships and how the desired code class structure can be generated.
Finally, Section~\ref{u_argtrees_gui} shows how a configuration file can be easily manipulated with a graphical user interface, allowing the user to create and change complex experiments without writing a single line of code.

\subsection{Modularity}
\label{u_argtrees_modularity}

As practiced in most non-simplistic codebases, the core of the argument tree structure is strong modularity.
The framework code is fragmented into different components with clearly defined purposes, such as training loops and datasets.
Exchanging modules of the same type for one another is a simple issue, for example gradient-descent optimizers.
If all implemented code classes of the same type inherit from one base class (e.g., AbstractOptimizer) that guarantees specific class methods for a stable interaction, they can be treated equally. In object-oriented programming, this design is termed polymorphism.

UniNAS extends typical PyTorch~\citep{pytorch} classes with additional functionality.
An example is image classification data sets, which ordinarily do not contain information about image sizes. Adding this specification makes it possible to use fake data easily and to precompute the tensor shapes in every layer throughout the neural network.

\begin{figure*}[ht]
\hfill
\begin{minipage}[c]{0.97\textwidth}
\begin{python}
@Register.task(search=True)
class SingleSearchTask(SingleTask):

	@classmethod
	def args_to_add(cls, index=None) -> [Argument]:
		return [
			Argument('is_test_run', default='False', type=str, is_bool=True),
			Argument('seed', default=0, type=int),`
			Argument('save_dir', default='{path_tmp}', type=str),
		]

	@classmethod
	def meta_args_to_add(cls) -> [MetaArgument]:
		methods = Register.methods.filter_match_all(search=True)
		return [
			MetaArgument('cls_device', Register.devices_managers, num=1),
			MetaArgument('cls_trainer', Register.trainers, num=1),
			MetaArgument('cls_method', methods, num=1),
		]
\end{python}
\end{minipage}
	\vskip-0.3cm
	\caption{
		UniNAS code excerpt for a SingleSearchTask. The decorator function in Line~1 registers the class with type ''task'' and additional information.
		The method in Line~5 returns all arguments for the task to be set in a config file.
		The method in Line~13 defines the local tree structure by stating how many modules of which types are needed. It is also possible to specify additional requirements, as done in Line~14.
	}
	\label{u_fig_register}
\end{figure*}

\subsection{A global register}
\label{u_argtrees_register}

A second requirement for argument trees is a global register for all modules. Its functions are:
\begin{itemize}
\setlength{\itemsep}{0pt}
\setlength{\parskip}{0pt}
\setlength{\parsep}{0pt}
	\item {
		Allow any module to register itself with additional information about its purpose. The example code in Figure~\ref{u_fig_register} shows this in Line~1.
	}
	\item {
		List all registered classes, including their type (task, model, optimizer, data set, and more) and their additional information (search, regression, and more).
	}
	\item {
		Filter registered classes by types and matching information.
	}
	\item {
		Given only the name of a registered module, return the class code located anywhere in the framework's files.
	}
\end{itemize}

As seen in the following Sections, this functionality is indispensable to UniNAS' design.
The only difficulties in building such a register is that the code should remain readable and that every module has to register itself when the framework is used.
Both can be achieved by scanning through all code files whenever a new job starts, which takes less than five seconds.
Python executes the decorators (see Figure~\ref{u_fig_register}, Line~1) by doing so, which handle registration in an easily readable fashion.


\subsection{Tree-based dependency structures}
\label{u_argtrees_tree}

\begin{figure*}
	\vskip-0.7cm
	\begin{minipage}[l]{0.42\linewidth}
		\centering
		\includegraphics[trim=0 320 2480 0, clip, width=\textwidth]{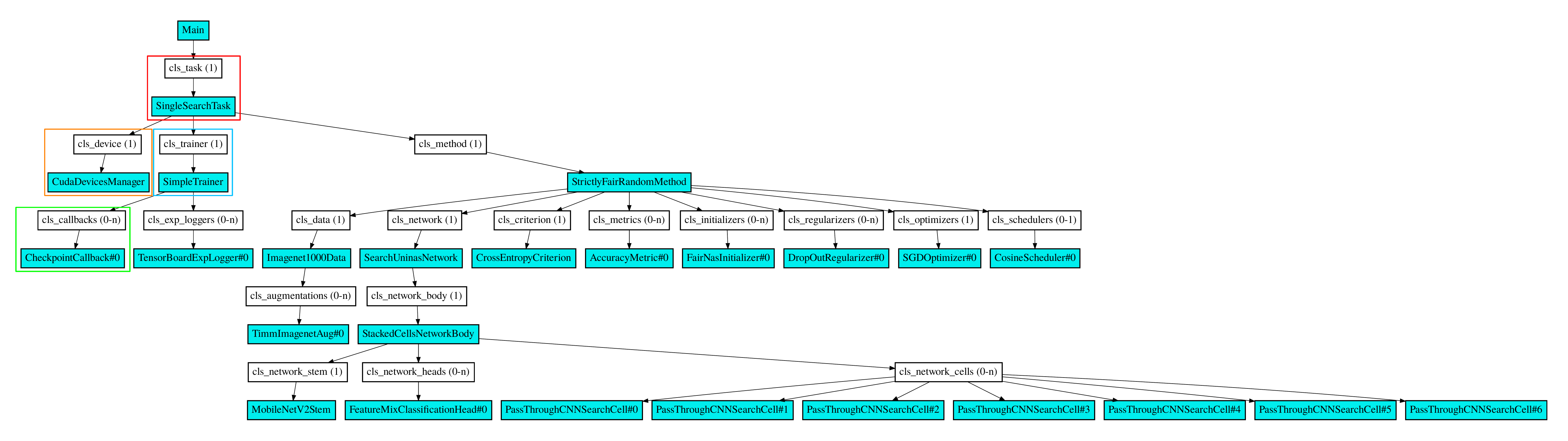}
		\vskip-0.2cm
		\caption{
			Part of a visualized SingleSearchTask configuration, which describes the training of a one-shot super-network with a specified search method (omitted for clarity, the complete tree is visualized in Figure~\ref{app_u_argstree_img}).
			The white colored tree nodes state the type and number of requested classes, the turquoise boxes the specific classes used. For example, the \textcolor{red}{SingleSearchTask} requires exactly one type of \textcolor{orange}{hardware device} to be specified, but the \textcolor{cyan}{SimpleTrainer} accepts any number of \textcolor{green}{callbacks} or loggers.
			\\
			\hfill
		}
		\label{u_argstree_trimmed_img}
	\end{minipage}
	\hfill
	\begin{minipage}[r]{0.5\textwidth}
		\begin{small}
\begin{lstlisting}[backgroundcolor = \color{white}]
"cls_task": <@\textcolor{red}{"SingleSearchTask"}@>,
"{cls_task}.save_dir": "{path_tmp}/",
"{cls_task}.seed": 0,
"{cls_task}.is_test_run": true,

"cls_device": <@\textcolor{orange}{"CudaDevicesManager"}@>,
"{cls_device}.num_devices": 1,

"cls_trainer": <@\textcolor{cyan}{"SimpleTrainer"}@>,
"{cls_trainer}.max_epochs": 3,
"{cls_trainer}.ema_decay": 0.5,
"{cls_trainer}.ema_device": "cpu",

"cls_exp_loggers": <@\textcolor{black}{"TensorBoardExpLogger"}@>,
"{cls_exp_loggers#0}.log_graph": false,

"cls_callbacks": <@\textcolor{green}{"CheckpointCallback"}@>,
"{cls_callbacks#0}.top_n": 1,
"{cls_callbacks#0}.key": "train/loss",
"{cls_callbacks#0}.minimize_key": true,
\end{lstlisting}
		\end{small}
		\vskip-0.2cm
		\caption{
			Example content of the configuration text-file (JSON format) for the tree in Figure~\ref{u_argstree_trimmed_img}.
			The first line in each text block specifies the used class(es), the other lines their detailed settings. For example, the \textcolor{cyan}{SimpleTrainer} is set to train for three epochs and track an exponential moving average of the network weights on the CPU.
		}
		\label{u_argstree_trimmed_text}
	\end{minipage}
\end{figure*}

A SingleSearchTask requires exactly one hardware device and exactly one training loop (named trainer, to train an over-complete super-network), which in turn may use any number of callbacks and logging mechanisms.
Their relationship is visualized in Figure~\ref{u_argstree_trimmed_img}.

Argument trees are extremely flexible since they allow every hierarchical one-to-any relationship imaginable.
Multiple optional callbacks can be rearranged in their order and configured in detail.
Moreover, module definitions can be reused in other constellations, including their requirements.
The ProfilingTask does not need a training loop to measure the runtime of different network topologies on a hardware device, reducing the argument tree in size.
While not implemented, a MultiSearchTask could use several trainers in parallel on several devices.

The hierarchical requirements  are made available using so-called MetaArguments, as seen in Line~16 of Figure~\ref{u_fig_register}.
They specify the local structure of argument trees by stating which other modules are required. To do so, writing the required module type and their amount is sufficient. As seen in Line~14, filtering the modules is also possible to allow only a specific subset.
This particular example defines the upper part of the tree visualized in Figure~\ref{u_argstree_trimmed_img}.
The names of all MetaArguments start with "cls\_" which improves readability and is reflected in the visualized arguments tree (Figure~\ref{u_argstree_trimmed_img}, white-colored boxes).

\subsection{Tree-based argument configurations}
\label{u_argtrees_config}

While it is possible to define such a dynamic structure, how can it be represented in a configuration file?
Figure~\ref{u_argstree_trimmed_text} presents an excerpt of the configuration that matches the tree in Figure~\ref{u_argstree_trimmed_img}.
As stated in Lines~6 and~9 of the configuration, CudaDevicesManager and SimpleTrainer fill the roles for the requested modules of types "device" and "trainer".
Lines~14 and~17 list one class of the types ''logger'' and ''callback'' each, but could provide any number of comma-separated names.
Also including the stated "task" type in Line~1, the mentioned lines state strictly which code classes are used and, given the knowledge about their hierarchy, define the tree structure.

Additionally, every class has some arguments (hyper-parameters) that can be modified.
SingleSearchTask defined three such arguments (Lines~7 to~9 in Figure~\ref{u_fig_register}) in the visualized example,
which are represented in the configuration (Lines~2 to~4 in Figure~\ref{u_argstree_trimmed_text}).
If the configuration is missing an argument, maybe to keep it short, its default value is used.
Another noteworthy mechanism in Line~2 is that "\{cls\_task\}.save\_dir" references whichever class is currently set as "cls\_task" (Line~1), without naming it explicitly.
Such wildcard references simplify automated changes to configuration files since, independently of the used task class, overwriting "\{cls\_task\}.save\_dir" is always an acceptable way to change the save directory.
A less general but perhaps more readable notation is "SingleSearchTask.save\_dir", which is also accepted here.

A very interesting property of such dynamic configuration files is that they contain only the hyper-parameters (arguments) of the used code classes.
Adding any additional arguments will result in an error since the configuration-parsing mechanism, described in Section~\ref{u_argtrees_build}, is then unable to piece the information together.
Even though UniNAS implements several different optimizer classes, any such configuration only contains the hyper-parameters of those used. Generated configuration files are always complete (contain all available arguments), sparse (contain only the available arguments), and never ambiguous.

A debatable design decision of the current configuration files, as seen in Figure~\ref{u_argstree_trimmed_text}, is that they do not explicitly encode any hierarchy levels. Since that information is already known from their class implementations, the flat representation was chosen primarily for readability.
It is also beneficial when arguments are manipulated, either automatically or from the terminal when starting a task.
The disadvantage is that the argument names for class types can only be used once ("cls\_device", "cls\_trainer", and more); an unambiguous assignment is otherwise not possible. For example, since the SingleSearchTask already owns "cls\_device", no other class that could be used in the same argument tree can use that particular name. While this limitation is not too significant, it can be mildly confusing at times.

Finally, how is it possible to create configuration files?
Since the dynamic tree-based approach offers a wide variety of possibilities, only a tiny subset is valid.
For example, providing two hardware devices violates the defined tree structure of a SingleSearchTask and results in a parsing failure.
If that happens, the user is provided with details of which particular arguments are missing or unexpected.
While the best way to create correct configurations is surely experience and familiarity with the code base, the same could be said about any framework.
Since UniNAS knows about all registered classes, which other (possibly specified) classes they use, and all of their arguments (including defaults, types, help string, and more), an exhaustive list can be generated automatically. However, resulting in almost 1600 lines of text, this solution is not optimal either.
The most convenient approach is presented in Section~\ref{u_argtrees_gui}: Creating and manipulating argument trees with a graphical user interface.

\begin{algorithm}
	\caption{
		Pseudo-code for building the argument tree, best understood with Figures~\ref{u_argstree_trimmed_img} and~\ref{u_argstree_trimmed_text}
		For a consistent terminology of code classes and tree nodes: If the $Task$ class uses a $Trainer$, then in that context, $Trainer$ the child. Lines starting with \# are comments.
	}
	\label{alg_u_argtree}
	\small
	\begin{algorithmic}
		\Require $Configuration$ \Comment{Content of the configuration file}
		\Require $Register$ \Comment{All modules in the code are registered}

		\State{}
		\State{$\#$ recursive parsing function to build a tree}
		\Function{parse}{$class,~index$}
		\Comment{E.g. $(SingleSearchTask,~0)$}

		\State $node = ArgumentTreeNode(class,~index)$

		\State{}
		\State{$\#$ first parse all arguments (hyper-parameters) of this tree node}

		\ForEach{($idx, argument\_name$) \textbf{in} $class.get\_arguments()$}
		\Comment{E.g. (0, $''save\_dir''$)}
		\State $value = get\_used\_value(Configuration,~class,~index,~argument\_name)$
		\State $node.add\_argument(argument\_name,~value)$
		\EndFor

		\State{}
		\State{$\#$ then recursively parse all child classes, for each module type...}

		\ForEach{$child\_class\_type$ \textbf{in} $class.get\_child\_types()$}
		\Comment{E.g. $cls\_trainer$}

		\State $class\_names = get\_used\_classes(Configuration,~child\_classes\_type)$

		\Assert{ The number of $class\_names$ is in the specified limits}
		\State{}
		\State{$\#$ for each module type, check all configured classes}

		\ForEach{($idx,~class\_name$) \textbf{in} $class\_names$}
		\Comment{E.g. (0, $''SimpleTrainer''$)}

		\State $child\_class = Register.get(child\_class\_name)$

		\State $child\_node = $\Call{parse}{$child\_class,~idx$}

		\State $node.add\_child(child\_class\_type,~idx,~child\_node)$
		\EndFor
		\EndFor

		\Returnx{ $node$}

		\EndFunction

		\State{}
		\State $tree = $\Call{parse}{$Main, 0$}
		\Comment{Recursively parse the tree, $Main$ is the entry point}

		\Ensure every argument in the configuration has been parsed
	\end{algorithmic}
\end{algorithm}

\subsection{Building the argument tree and code structure}
\label{u_argtrees_build}

The arguably most important function of a research code base is to run experiments.
In order to do so, valid configuration files must be translated into their respective code structure.
This comes with three major requirements:

\begin{itemize}
\setlength{\itemsep}{0pt}
\setlength{\parskip}{0pt}
\setlength{\parsep}{0pt}
	\item{
		Classes in the code that implement the desired functionality.
		As seen in Section~\ref{u_argtrees_tree} and Figure~\ref{u_argstree_trimmed_img}, each class also states the types, argument names and numbers of additionally requested classes for the local tree structure.
	}
	\item{
		A configuration that describes which code classes are used and which values their parameters take.
		This is described in Section~\ref{u_argtrees_config} and visualized in Figure~\ref{u_argstree_trimmed_text}.
	}
	\item{
		To connect the configuration content to classes in the code, it is required to reference code modules by their names. As described in Section~\ref{u_argtrees_register} this can be achieved with a global register.
	}
\end{itemize}

Algorithm~\ref{alg_u_argtree} realizes the first step of this process: parsing the hierarchical code structure and their arguments from the flat configuration file.
The result is a tree of \textit{ArgumentTreeNodes}, of which each refers to exactly one class in the code, is connected to all related tree nodes, and knows all relevant hyper-parameter values.
While they do not yet have actual class instances, this final step is no longer difficult.

\begin{figure*}[h]
	\vskip -0.0in
	\begin{center}
		\includegraphics[trim=30 180 180 165, clip, width=\linewidth]{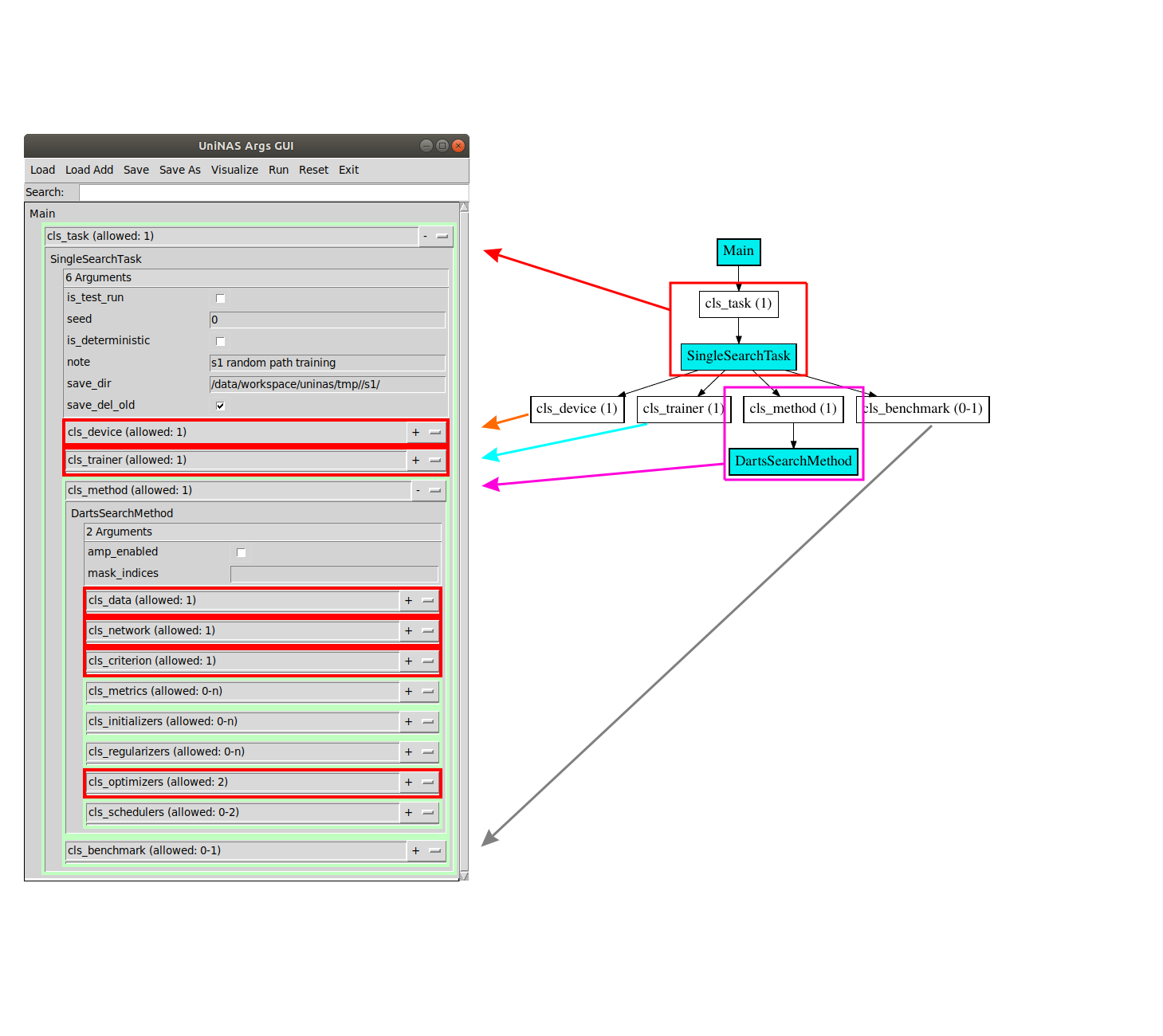}
		\hspace{-0.5cm}
		\caption{
			The graphical user interface (left) that can manipulate the configurations of argument trees (visualized right).
			Since many nodes are missing classes of some type ("cls\_device", ...), their parts in the GUI are highlighted in red.
			The eight child nodes of DartsSearchMethod are omitted for visual clarity.
		}
		\label{fig_u_gui}
	\end{center}
\end{figure*}

\subsection{Creating and manipulating argument trees with a GUI}
\label{u_argtrees_gui}

Manually writing a configuration file can be perplexing since one must keep track of tree specifications, argument names, available classes, and more.
The graphical user interface (GUI) visualized in Figures~\ref{fig_u_gui} and~\ref{app_u_gui} solves these problems to a large extent, by providing the following functionality:

\begin{itemize}
\setlength{\itemsep}{0pt}
\setlength{\parskip}{0pt}
\setlength{\parsep}{0pt}
	\item{
		Interactively add and remove nodes in the argument tree, thus also in the configuration and class structure. Highlight violations of the tree specification.
	}
	\item{
		Setting the hyper-parameters of each node, using checkboxes (boolean), dropdown menus (choice from a selection), and text fields (other cases like strings or numbers) where appropriate.
	}
	\item{
		Functions to save and load argument trees.
		Since it makes sense to separate the configurations for the training procedure and the network design to swap between different constellations easily, loading partial trees is also supported. Additional functions enable visualizing, resetting, and running the current argument tree.
	}
	\item{
		A search function that highlights all matches since the size of some argument trees can make finding specific arguments tedious.
	}
\end{itemize}

In order to do so, the GUI manipulates \textit{ArgumentTreeNodes} (Section~\ref{u_argtrees_build}), which can be easily converted into configuration files and code.
As long as the required classes (for example, the data set) are already implemented, the GUI enables creating and changing experiments without ever touching any code or configuration files.
While not among the original intentions, this property may be especially interesting for non-programmers that want to solve their problems quickly.

Still, the current version of the GUI is a proof of concept.
It favors functionality over design, written with the plain Python Tkinter GUI framework and based on little previous GUI programming experience.
Nonetheless, since the GUI (frontend) and the functions manipulating the argument tree (backend) are separated, a continued development with different frontend frameworks is entirely possible.
The perhaps most interesting would be a web service that runs experiments on a server, remotely configurable from any web browser.

\subsection{Using external code}
\label{u_external}

There is a variety of reasons why it makes sense to include external code into a framework.
Most importantly, the code either solves a standing problem or provides the users with additional options. Unlike newly written code, many popular libraries are also thoroughly optimized, reviewed, and empirically validated.

External code is also a perfect match for a framework based on argument trees.
As shown in Figure~\ref{u_fig_external_import}, external classes of interest can be thinly wrapped to ensure compatibility, register the module, and specify all hyper-parameters for the argument tree.
The integration is seamless so that finding out whether a module is locally written or external requires an inspection of its code.
On the other hand, if importing the AdaBelief~\citep{zhuang2020adabelief} code fails, the module will not be registered and therefore not be available in the graphical user interface.
UniNAS fails to parse configurations that require unregistered modules but informs the user which external sources can be installed to extend its functionality.

Due to this logistic simplicity, several external frameworks extend the core of UniNAS.
Some of the most important ones are:
\begin{itemize}
\setlength{\itemsep}{0pt}
\setlength{\parskip}{0pt}
\setlength{\parsep}{0pt}
	\item{
		pymoo~\citep{pymoo}, a library for multi-objective optimization methods.
	}
	\item{
		Scikit-learn~\citep{sklearn}, which implements many classical machine learning algorithms such as Support Vector Machines and Random Forests.
	}
	\item{
		PyTorch Image Models~\citep{rw2019timm}, which provides the code for several optimizers, network models, and data augmentation methods.
	}
	\item{
		albumentations~\citep{2018arXiv180906839B}, a library for image augmentations.
	}
\end{itemize}

\begin{figure*}
\hfill
\begin{minipage}[c]{0.95\textwidth}
\begin{python}
from uninas.register import Register
from uninas.training.optimizers.abstract import WrappedOptimizer
try:
    from adabelief_pytorch import AdaBelief
   	# if the import was successful,
   	# register the wrapped optimizer
    @Register.optimizer()
    class AdaBeliefOptimizer(WrappedOptimizer):
    	# wrap the original
    	...

except ImportError as e:
	# if the import failed,
	# inform the user that optional libraries are not installed
	Register.missing_import(e)
\end{python}
\end{minipage}
	\vskip-0.3cm
	\caption{
		Excerpt of UniNAS wrapping the official AdaBelief optimizer code.
		The complete text has just 45 lines, half of which specify the optimizer parameters for the argument trees.
	}
	\label{u_fig_external_import}
\end{figure*}

\section{Dynamic network designs}
\label{u_networks}

As seen in the previous Sections, the unique design of UniNAS enables powerful customization of all components. In most cases, a significant portion of the architecture search configuration belongs to the network design. The FairNAS search example in Figure~\ref{app_u_argstree_img} contains 25 configured classes, of which 11 belong to the search network.
While it would be easy to create a single configurable class for each network architecture of interest, that would ignore the advantages of argument trees.
On the other hand, there are many technical difficulties with highly dynamic network topologies. Some of them are detailed below.

\subsection{Decoupling components}

In many published research codebases, network and architecture weights jointly exist in the network class. This design decision is disadvantageous for multiple reasons.
Most importantly, changing the network or NAS method requires a lot of manual work.
The reason is that different NAS methods need different amounts of architecture parameters, use them differently, and optimize them in different ways. For example:
\begin{itemize}[noitemsep,parsep=0pt,partopsep=0pt]
	\item{
		DARTS~\citep{liu2018darts} requires one weight vector per architecture choice.
		They weigh all different paths, candidate operations, in a sum. Updating the weights is done with an additional optimizer (ADAM), using gradient descent.
	}
	\item{
		MDENAS~\citep{mdenas} uses a similar vector for a weighted sample of a single candidate operation that is used in this particular forward pass. Global network performance feedback is used to increase or decrease the local weightings.
	}
	\item{
		Single-Path One-Shot~\citep{guo2020single} does not use weights at all. Paths are always sampled uniformly randomly. The trained network is used as an accuracy prediction model and used by a hyper-parameter optimization method.
	}
	\item{
		FairNAS~\citep{FairNAS} extends Single-Path One-Shot to make sure that all candidate operations are used frequently and equally often. It thus needs to track which paths are currently available.
	}
\end{itemize}

\begin{figure}[t]
	\vskip -0.0in
	\begin{center}
		\includegraphics[trim=0 0 0 0, clip, width=\linewidth]{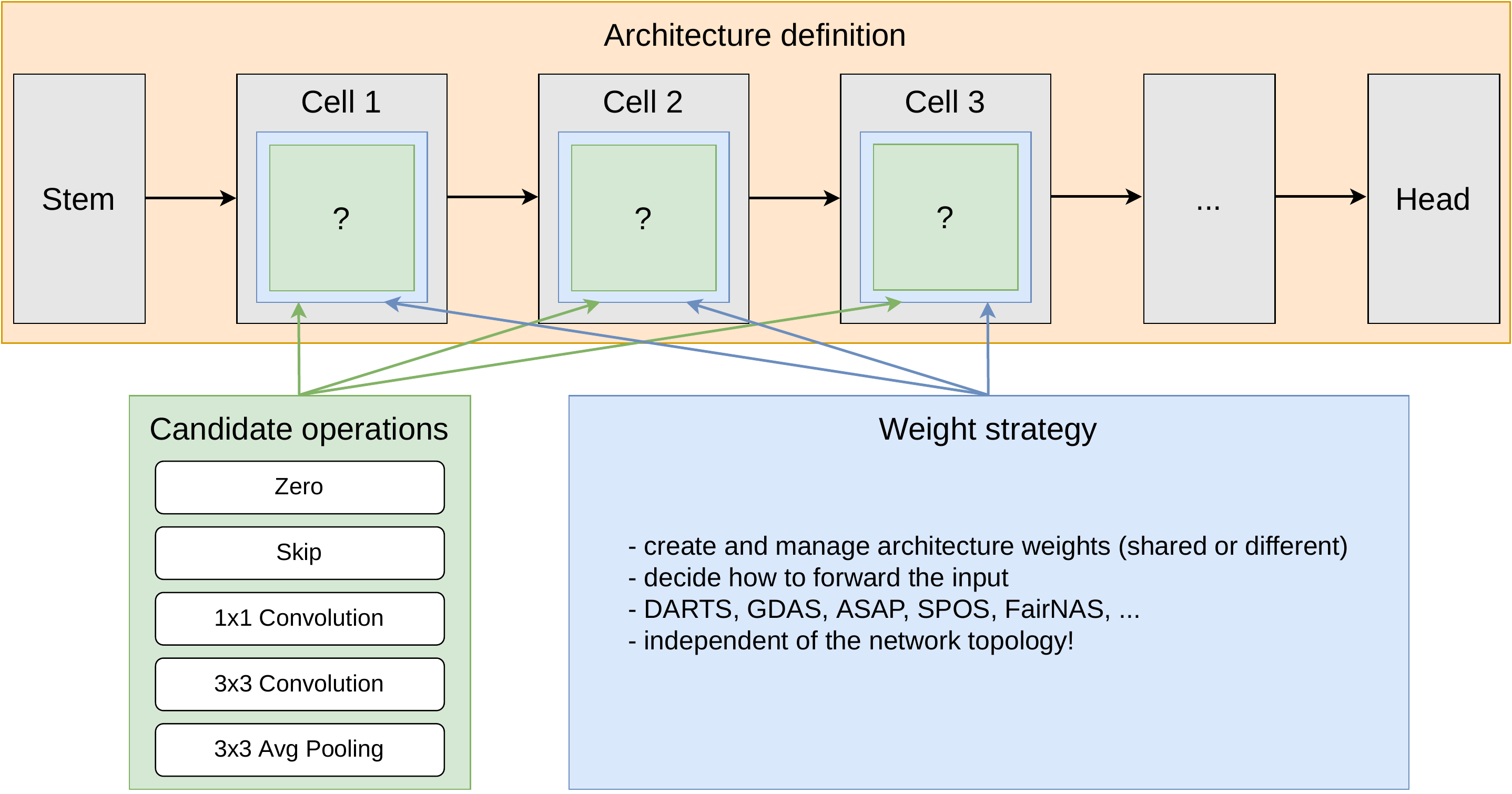}
		\hspace{-0.5cm}
		\caption{
			The network and architecture weights are decoupled.
			\textbf{Top}: The structure of a fully sequential super-network. Every layer (cell) uses the same set of candidate operations and weight strategy.
			\textbf{Bottom left}: One set of candidate operations that is used multiple times in the network. This particular experiment uses the NAS-Bench-201 candidate operations.
			\textbf{Bottom right}: A weight strategy that manages everything related to the used NAS method, such as creating the architecture weights or which candidates are used in each forward pass.
		}
		\label{fig_u_decouple}
	\end{center}
\end{figure}

The same is also true for the set of candidate operations, which affect the sizes of the architecture weights.
Once the definitions of the search space, the candidate operations, and the NAS method (including the architecture weights) are mixed, changing any part is tedious.
Therefore, strictly separating them is the best long-term approach.
Similar to other frameworks presented in Section~\ref{u_introduction_available},
architectures defined in UniNAS do not use an explicit set of candidate architectures but allow a dynamic configuration.
This is supported by a \textit{WeightStrategy} interface, which handles all NAS-related operations such as creating and updating the architecture weights.
The interaction between the architecture definition, the candidate operations, and the weight strategy is visualized in Figure~\ref{fig_u_decouple}.

The easy exchange of any component is not the only advantage of this design.
Some NAS methods, such as DARTS, update network and architecture weights using different gradient descent optimizers. Correctly disentangling the weights is trivial if they are already organized in decoupled structures but hard otherwise.
Another advantage is that standardizing functions to create and manage architecture weights makes it easy to present relevant information to the user, such as how many architecture weights exist, their sizes, and which are shared across different network cells.
An example is presented in Figure~\ref{app_text}.

\begin{figure}[hb!]
\begin{minipage}[c]{0.24\textwidth}
	\centering
	\includegraphics[height=11.5cm]{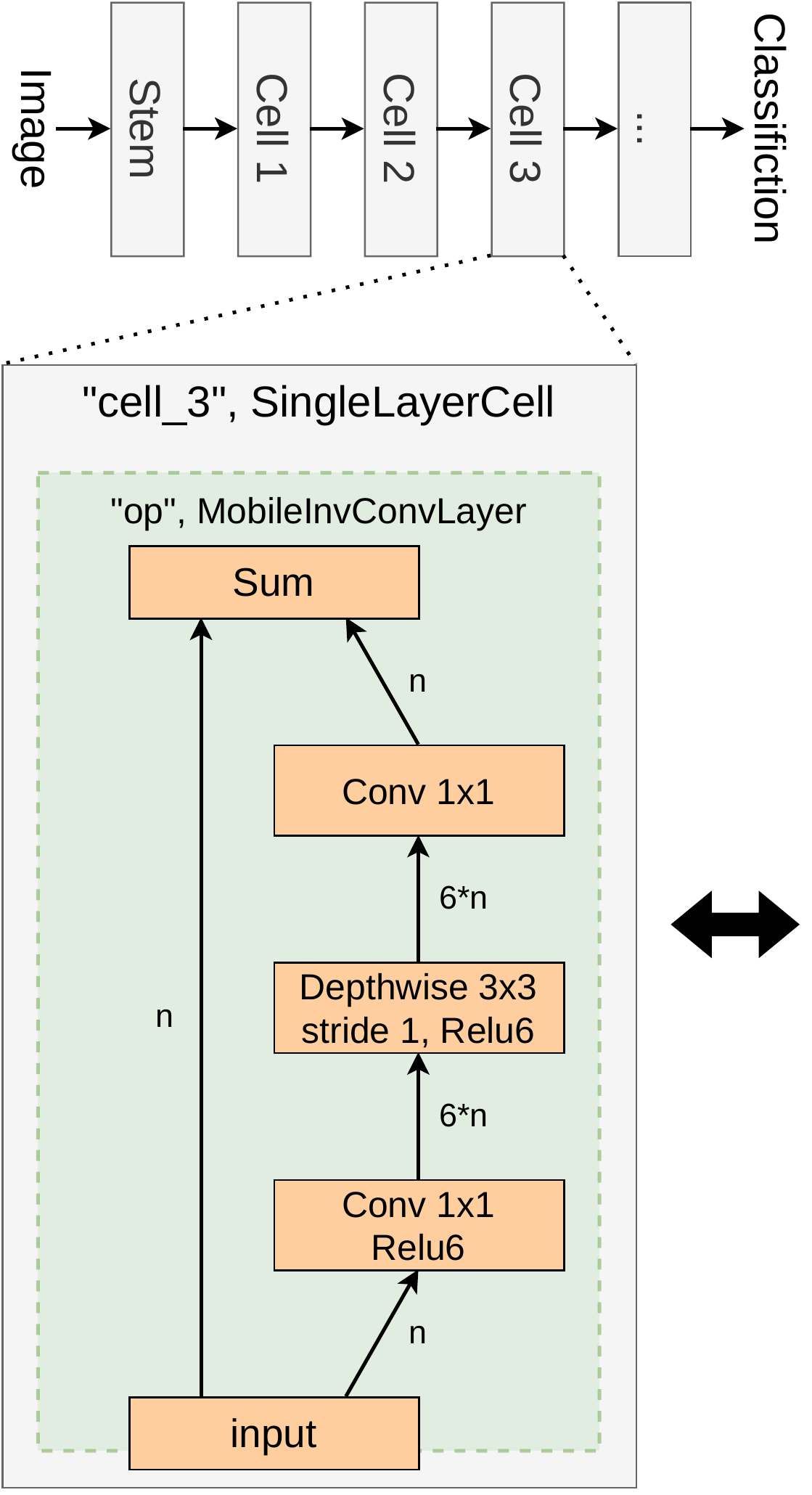}
\end{minipage}
\hfill
\begin{minipage}[c]{0.5\textwidth}
\small
\begin{python}
"cell_3": {
	"name": "SingleLayerCell",
	"kwargs": {
		"name": "cell_3",
		"features_mult": 1,
		"features_fixed": -1
	},
	"submodules": {
		"op": {
			"name": "MobileInvConvLayer",
			"kwargs": {
				"kernel_size": 3,
				"kernel_size_in": 1,
				"kernel_size_out": 1,
				"stride": 1,
				"expansion": 6.0,
				"padding": "same",
				"dilation": 1,
				"bn_affine": true,
				"act_fun": "relu6",
				"act_inplace": true,
				"att_dict": null,
				"fused": false
			}
		}
	}
},
\end{python}
\end{minipage}
	\caption{
		 A high-level view on the MobileNet~V2 architecture~\citep{sandler2018mobilenetv2} in the top left,
		 and a schematic of the inverted bottleneck block in the bottom left.
		 This design uses two 1$\times$1 convolutions to change the channel count \textit{n} by an expansion factor of~6, and a spatial 3$\times$3 convolution in their middle.
		 The text on the right-hand side represents the cell structure by referencing the modules by their names ("name") and their keyworded arguments ("kwargs").
	}
	\label{u_fig_conf}
\end{figure}

\subsection{Saving, loading, and finalizing networks}
\label{u_networks_save}

As mentioned before, argument trees enable a detailed configuration of every aspect of an experiment, including the network topology itself.
As visualized in Figure~\ref{app_u_argstree_img}, such network definitions can become almost arbitrarily complex.
This becomes disadvantageous once models have to be saved or loaded or when super-networks are finalized into discrete architectures.
Unlike TensorFlow~\citep{tensorflow2015-whitepaper}, the used PyTorch~\citep{pytorch} library saves only the network weights without execution graphs.
External projects like ONNX~\citep{onnx} can be used to export limited graph information but not to rebuild networks using the same code classes and context.

The implemented solution is inspired by the official code\footnote{\url{https://github.com/mit-han-lab/proxylessnas/tree/master/proxyless_nas}} of ProxylessNAS~\citep{proxylessnas}, where every code module defines two functions that enable exporting and importing the entire module state and context.
As typical for hierarchical structures, the state of an outer module contains the states of all modules within.
An example is visualized in Figure~\ref{u_fig_conf}, where one cell in the famous MobileNet V2 architecture is represented as readable text.
The global register can provide any class definition by name (see Section~\ref{u_argtrees_register}) so that an identical class structure can be created and parameterized accordingly.

The same approach that enables saving and loading arbitrary class compositions can also be used to change their structure.
More specifically, an over-complete super-network containing all possible candidate operations can export only a specific configuration subset. The network recreated from this reduced configuration is the result of the architecture search.
This is made possible since the weight strategy controls the use of all candidate operations, as visualized in Figure~\ref{fig_u_decouple}. Similarly, when their configuration is exported, the weight strategy controls which candidates should be part of the finalized network architecture.
In another use case, some modules behave differently in super-networks and finalized architectures. For example, Linear Transformers~\citep{ScarletNAS} supplement skip connections with linear 1$\times$1 convolutions in super-networks to stabilize the training with variable network depths.
When the network topology is finalized, it suffices to simply export the configuration of a skip connection instead of their own.

Another practical way of rebuilding code structures is available through the argument tree configuration, which defines every detail of an experiment (see Section~\ref{u_argtrees_config}).
Parsing the network design and loading the trained weights of a previous experiment requires no further user interaction than specifying its save directory.
This specific way of recreating experiment environments is used extensively in \textit{Single-Path One-Shot} tasks.
In the first step, a super-network is trained to completion. Afterward, when the super-network is used to make predictions for a hyper-parameter optimization method (such as Bayesian optimization or evolutionary algorithms), the entire environment of its training can be recreated. This includes the network design and the dataset, data augmentations, which parts were reserved for validation, regularization techniques, and more.

\section{Discussion and Conclusions}
\label{u_conclusions}

We presented the underlying concepts of UniNAS, a PyTorch-based framework with the ambitious goal of unifying a variety of NAS algorithms in one codebase.
Even though the use cases for this framework changed over time, mostly from DARTS-based to SPOS-based experiments, its underlying design approach made reusing old code possible at every step.
However, several technical details could be changed or improved in hindsight. Most importantly, configuration files should reflect the hierarchy levels (see Section~\ref{u_argtrees_config}) for code simplicity and to avoid concerns about using module types multiple times. The current design favors readability, which is now a minor concern thanks to the graphical user interface.
Other considered changes would improve the code readability but were not implemented due to a lack of necessity and time.

In summary, the design of UniNAS fulfills all original requirements.
Modules can be arranged and combined in almost arbitrary constellations, giving the user an extremely flexible tool to design experiments.
Furthermore, using the graphical user interface does not require writing even a single line of code.
The resulting configuration files contain only the relevant information and do not suffer from a framework with many options.
These features also enable an almost arbitrary network design, combined with any NAS optimization method and any set of candidate operations. Despite that, networks can still be saved, loaded, and changed in various ways.
Although not covered here, several unit tests ensure that the essential framework components keep working as intended.

Finally, what is the advantage of using argument trees over writing code with the same results?
Compared to configuration files, code is more powerful and versatile but will likely suffer from problems described in Section~\ref{u_introduction_available}.
Argument trees make any considerations about which parameters to expose unnecessary and can enforce the use of specific module types and subsets thereof.
However, their strongest advantage is the visualization and manipulation of the entire experiment design with a graphical user interface. This aligns well with Automated Machine Learning (AutoML), which is also intended to make machine learning available to a broader audience.

{\small
\bibliographystyle{iclr2022_conference}
\bibliography{./bib}  
}

\newpage

\appendix

\section{Additional resources}
\label{app_images}

\begin{figure*}[h]
	\begin{tiny}

\begin{verbatim}
>---------------------------------------------- Args ----------------------------------------------<
 > cls_task                                  SingleSearchTask                                task
 > cls_device                                CudaDevicesManager                              device manager
 > cls_trainer                               SimpleTrainer                                   trainer
 > cls_method                                UniformRandomMethod                             method
 > cls_benchmark                                                                             immediately look up the search result in this benchmark set (optional)
 > cls_callbacks                             CheckpointCallback                              training callbacks
 > cls_clones                                                                                training clones
 > cls_exp_loggers                           TensorBoardExpLogger                            experiment logger
 > cls_data                                  Cifar10Data                                     data set
 > cls_network                               SearchUninasNetwork                             network
 > cls_criterion                             CrossEntropyCriterion                           criterion
 > cls_metrics                               AccuracyMetric                                  training metric
 > cls_initializers                                                                          weight initializer
 > cls_regularizers                          DropOutRegularizer                              regularizer
 > cls_optimizers                            SGDOptimizer                                    optimizer
 > cls_schedulers                            CosineScheduler                                 scheduler
 > cls_augmentations                         DartsCifarAug                                   data augmentation
 > cls_network_body                          StackedCellsNetworkBody                         network
 > cls_network_stem                          ConvStem                                        network stem
 > cls_network_heads                         Bench201Head                                    network heads
 > cls_network_cells                         Bench201CNNSearchCell, Bench201ReductionCell    network cells
 > cls_network_cells_primitives              Bench201Primitives, Bench201Primitives          network cells primitives
 > SingleSearchTask.is_test_run              True                                            test runs stop epochs early
 > SingleSearchTask.seed                     0                                               random seed for the experiment
 > SingleSearchTask.is_deterministic         False                                           use deterministic operations
 > SingleSearchTask.note                     s1 SPOS-like training                           just to take notes
 > SingleSearchTask.save_dir                 /tmp/demo/icw/train_supernet/                   where to save
 > SingleSearchTask.save_del_old             True                                            wipe the save dir before starting
 > CudaDevicesManager.num_devices            1                                               number of available devices
 > CudaDevicesManager.use_cudnn              True                                            try using cudnn
 > CudaDevicesManager.use_cudnn_benchmark    True                                            use cudnn benchmark
 > SimpleTrainer.max_epochs                  10                                              max training epochs, affects schedulers + regularizers
 > SimpleTrainer.stop_epoch                  -1                                              stop after training n epochs anyway, if > 0
 > SimpleTrainer.log_fs                      False                                           log file system usage
 > SimpleTrainer.log_ram                     False                                           log RAM usage
 > SimpleTrainer.log_device                  True                                            log device usage
 > SimpleTrainer.eval_last                   10                                              run eval for the last n epochs, always if <0
 > SimpleTrainer.test_last                   10                                              run test for the last n epochs, always if <0
 > SimpleTrainer.accumulate_batches          1                                               accumulate gradients over n batches before stepping updating. Does not change the learning rate, may cause issues when there are multiple alternating optimizers
...
 > StackedCellsNetworkBody.cell_order        n, n, r, n, n, r, n, n                          arrangement of cells
 > ConvStem.features                         16                                              num output features of this stem
...

>----------------------------------------- setting up... ------------------------------------------<
Data Set: splitting the training set, will use 5000 data points as validation set
Building StackedCellsNetworkBody:
    cell index          name    class              input shapes               output shapes          #params
                        -       ConvStem           Shape(3, 32, 32)           [Shape(16, 32, 32)]    464
    0                   n       Bench201CNNCell    [Shape(16, 32, 32)]        [Shape(16, 32, 32)]    18160
    1                   n       Bench201CNNCell    [Shape(16, 32, 32)]        [Shape(16, 32, 32)]    18160
    2                   r       SingleLayerCell    [Shape(16, 32, 32)]        [Shape(32, 16, 16)]    14464
    3                   n       Bench201CNNCell    [Shape(32, 16, 16)]        [Shape(32, 16, 16)]    67040
    4                   n       Bench201CNNCell    [Shape(32, 16, 16)]        [Shape(32, 16, 16)]    67040
    5                   r       SingleLayerCell    [Shape(32, 16, 16)]        [Shape(64, 8, 8)]      57600
    6                   n       Bench201CNNCell    [Shape(64, 8, 8)]          [Shape(64, 8, 8)]      256960
    7                   n       Bench201CNNCell    [Shape(64, 8, 8)]          [Shape(64, 8, 8)]      256960
                        -       Bench201Head       Shape(64, 8, 8)            Shape(10)              778
    complete network                               Shape(3, 32, 32)           [Shape(10)]            757626
Network built, it has 757626 parameters!
Using device: CudaDeviceMover([0])
Continuously logging (devices=CudaDeviceMover([0]), RAM=False, file_system=False) each 5s
...

>---------------------------------------- Weight strategy -----------------------------------------<
RandomChoiceStrategy("default", 6 architecture weights)
Weights:
   name                num choices    used
 > n/block-0/1/op-0    5              6x
 > n/block-1/2/op-0    5              6x
 > n/block-1/2/op-1    5              6x
 > n/block-2/3/op-0    5              6x
 > n/block-2/3/op-1    5              6x
 > n/block-2/3/op-2    5              6x
\end{verbatim}
	\end{tiny}
	\vskip-0.2cm
	\caption{
		Excerpts of UniNAS' text output.
		\textbf{Top}: The names, values, and help text of all (meta-) arguments. The effect of the last two can be observed in the network structure.
		\textbf{Center}: Since the network code is well-defined, it is possible to generate an overview of layers, inputs, outputs, and parameters.
		\textbf{Bottom}: The weight strategy can present the interesting information about the used architecture weights. There are five candidates in the chosen operation set (\textit{Bench201Primitives}), and six cells "n" with shared architecture.
	}
	\label{app_text}
\end{figure*}

\begin{figure*}[h]
	\vskip -0.0in
	\begin{center}
		\includegraphics[trim=40 0 10 0, clip, height=10cm]{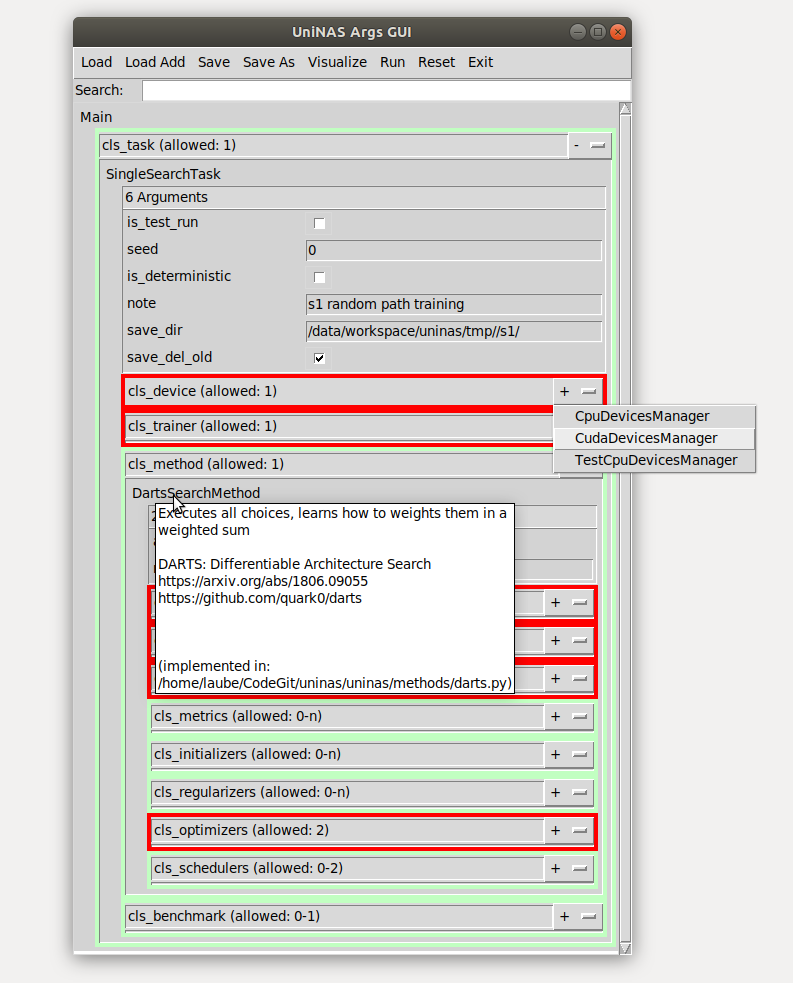}
		\hfill
		\includegraphics[trim=20 0 40 0, clip, height=10cm]{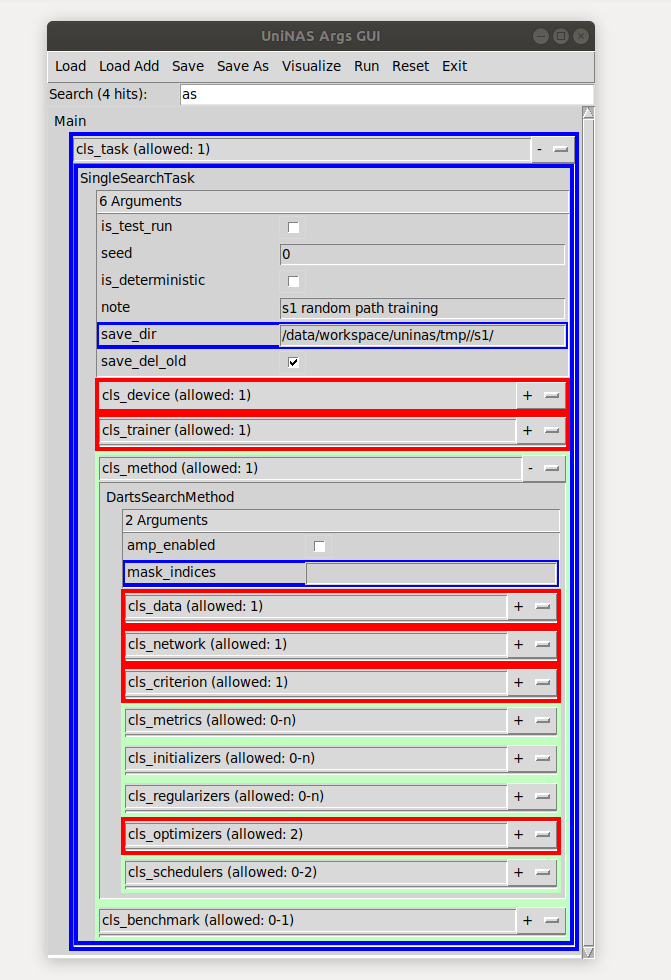}

		\caption{
			Additional images for the graphical user interface (GUI).\\
			\textbf{Left}: Hovering the mouse cursor over any name brings up a tooltip, describing the comment in the code and in which file it is implemented. Pressing the Plus and Minus dropdown buttons on the right side enables adding and removing any appropriate classes in the tree structure.\\
			\textbf{Right}: By adding a search text (top), matches are highlighted in blue. The text "as" can be present in argument names ("cls\_t\textbf{as}k", "m\textbf{as}k\_indices"), module names ("SingleSearchT\textbf{as}k"), or argument values ("save\_dir" has ".../unin\textbf{as}/...").
		}
		\label{app_u_gui}
	\end{center}
\end{figure*}

\begin{sidewaysfigure}
	\includegraphics[trim=0 0 0 0, clip, width=1\textwidth]{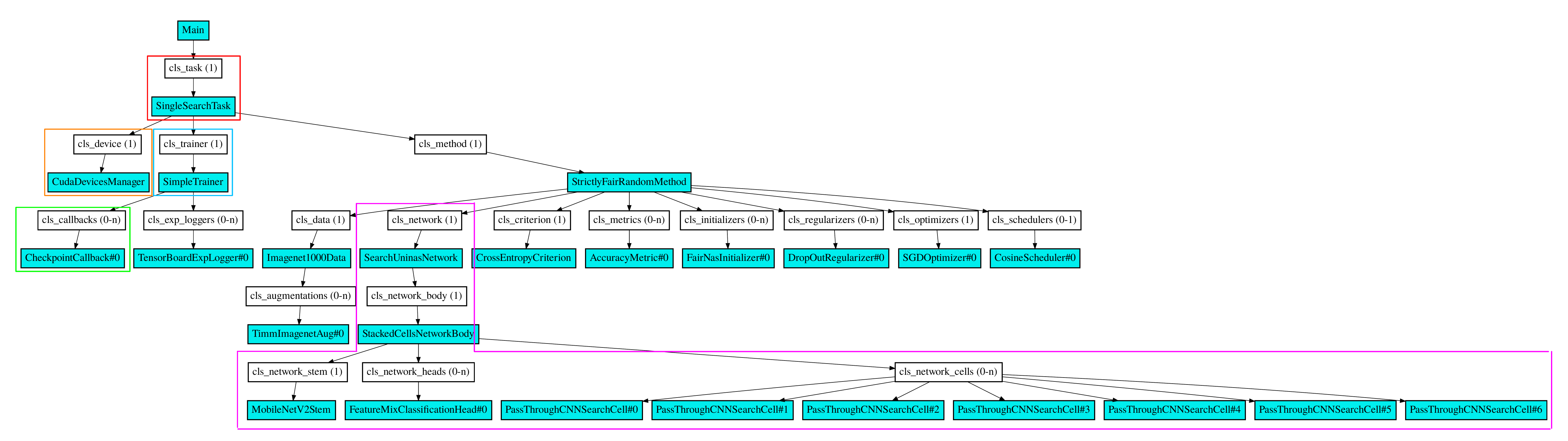}
	\caption{
		The full argument tree to train a FairNAS-like super-network, extending Figure~\ref{u_argstree_trimmed_img}.
		The model is trained on ImageNet using the CrossEntropy loss, SGD and a cosine learning rate schedule. The tracked accuracy is used for checkpointing, and dropout is enabled.
		The different cell definition model has six stages with defined numbers of channels (32, 40, 80, 96, 192, 320), the last cell definition retains the current input and output sizes.
	}
	\label{app_u_argstree_img}
\end{sidewaysfigure}
\newpage

\begin{figure*}
\hfill
\begin{minipage}[c]{0.95\textwidth}
\begin{python}
@Register.network_mixed_op()
class MixedOp(SumParallelModules):
	def __init__(self, submodules: list, name, strategy_name):
		# store all arguments, thus including them in the config
		super().__init__(submodules)
		self._add_to_kwargs(name=name, strategy_name=strategy_name)

		# create the needed architecture weights
		self.sm = StrategyManager()  # singleton class
		self.ws = self.sm.make_weight(strategy_name, name, submodules)

	def forward(self, x: torch.Tensor) -> torch.Tensor:
		# let the weight strategy decide how to forward inputs
		return self.ws.combine(self.name, x, self.submodules)

	def config(self, finalize=True, **_) -> dict:
		# describe this module, so that it can be rebuilt later
		if finalize:
			# only a subset of the candidates are requested
			# ask the weight strategy which candidates are best
			indices = self.ws.get_finalized_indices(self.name)
			modules = [self.submodules[i] for i in indices]
			if len(modules) == 1:
				return modules[0].config(finalize, **_)
			return SumParallelModules(modules).config(finalize, **_)
		else:
			# the entire super-network is requested
			return super().config(finalize=finalize, **_)

	@classmethod
	def from_config(cls, **kwargs):
		# the rebuilding of owned code sub-modules is omitted
		# the global register is used to create Modules by name
		submodules_ = ...
		submodule_lists_ = ...
		submodule_dicts_ = ...
		# rebuild this module with the exact same arguments as before
		return cls(**submodules_, **submodule_lists_, **submodule_dicts_, **kwargs)
\end{python}
\end{minipage}
	\caption{
		Excerpt of the UniNAS MixedOp code, an operation that manages multiple candidate operations.
		They are stored (Lines~5 and~6) and used in a forward pass (Line~14).
		The methods starting in Lines~16 and~31 define how this MixedOp module is exported as a JSON description and later rebuilt from such.
		The \textit{from\_config} function belongs to a super-class that every UniNAS network module inherits from and is not required to be implemented again in any new class. It is only displayed for completeness.
	}
	\label{app_u_fig_mixedop}
\end{figure*}

\end{document}